\documentclass[runningheads]{llncs}

\usepackage{amsmath,amsfonts,bm}

\def\eqref#1{equation~\ref{#1}}

\def\1{\bm{1}}

\def\vb{{\bm{b}}}

\def\vs{{\bm{s}}}

\def\vu{{\bm{u}}}

\def\vx{{\bm{x}}}

\DeclareMathAlphabet{\mathsfit}{\encodingdefault}{\sfdefault}{m}{sl}
\SetMathAlphabet{\mathsfit}{bold}{\encodingdefault}{\sfdefault}{bx}{n}

\newcommand{\E}{\mathbb{E}}

 \usepackage{graphicx}
\usepackage{hyperref}
\usepackage{url}

\usepackage[ruled,linesnumbered]{algorithm2e}

\usepackage{amsmath,amssymb}
\usepackage{thmtools}
\usepackage{graphicx}
\usepackage{subfigure}
\graphicspath{{imgs/}}
\usepackage{enumitem}
\usepackage{xcolor}

\usepackage{diagbox}
\usepackage{makecell}

\usepackage[firstpage]{draftwatermark}

\begin{document}

\title{Probabilistic Counterexample Guidance for Safer Reinforcement Learning (Extended Version)}

\titlerunning{Probabilistic Counterexample Guidance for Safer RL}
\author{Xiaotong Ji \and
Antonio Filieri}
\authorrunning{Ji and Filieri}
\institute{Department of Computing\\
Imperial College London\\
London, SW7 2AZ, UK \\
\email{\{xiaotong.ji16, a.filieri\}@imperial.ac.uk}\\}
\maketitle              \begin{abstract}
Safe exploration aims at addressing the limitations of Reinforcement Learning (RL) in safety-critical scenarios, where failures during trial-and-error learning may incur high costs. Several methods exist to incorporate external knowledge or to use proximal sensor data to limit the exploration of unsafe states. However, reducing exploration risks in unknown environments, where an agent must discover safety threats during exploration, remains challenging.

In this paper, we target the problem of safe exploration by guiding the training with counterexamples of the safety requirement. Our method abstracts both continuous and discrete state-space systems into compact abstract models representing the safety-relevant knowledge acquired by the agent during exploration. We then exploit probabilistic counterexample generation to construct minimal simulation submodels eliciting safety requirement violations, where the agent can efficiently train offline to refine its policy towards minimising the risk of safety violations during the subsequent online exploration.

We demonstrate our method’s effectiveness in reducing safety violations during online exploration in preliminary experiments by an average of 40.3\% compared with QL and DQN standard algorithms and 29.1\% compared with previous related work, while achieving comparable cumulative rewards with respect to unrestricted exploration and alternative approaches.

\keywords{Safe reinforcement learning \and Probabilistic model checking \and Counterexample guidance.}
\end{abstract}

\SetWatermarkText{\hspace*{11cm}\raisebox{14cm}{\includegraphics{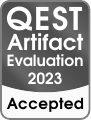}}}
\SetWatermarkAngle{0}
\section{Introduction}\label{secIntroduction}
A critical limitation of applying Reinforcement Learning (RL) in real-world control systems is its lack of guarantees of avoiding unsafe behaviours. At its core, RL is a trial-and-error process, where the learning agent explores the decision space and receives rewards for the outcome of its decisions. However, in safety-critical scenarios, failing trials may result in high costs or unsafe situations and should be avoided as much as possible. 

Several learning methods try to incorporate the advantages of model-driven and data-driven methods to encourage safety during learning~\cite{kim2020safe,garcia2015comprehensive}. One natural approach for encouraging safer learning is to analyse the kinematic model of the learning system with specific safety requirements and to design safe exploration~\cite{garcia2012safe,alshiekh2018safe,pham2018optlayer} or safe optimisation~\cite{achiam2017constrained,tessler2018reward,stooke2020responsive} strategies that avoid unsafe states or minimise the expected occurrence of unsafe events during training. However, this approach is not applicable for most control systems with partially-known or unknown dynamics, where not enough information is available to characterise unsafe states or events \textit{a priori}. 

To increase the safety of learning in environments with (entirely or partially) unknown dynamics, we propose an online-offline learning scheme where online execution traces collected during exploration are used to construct an abstract representation of the visited state-action space. If during exploration the agent violates a safety requirement with unacceptable frequency, a probabilistic model checker is used to produce from the abstract representation minimal counterexample sub-models, i.e., a minimal subset of the abstract state-action space within which the agent is expected to violate its safety requirement with a probability larger than tolerable. These counterexamples are then used to synthesise small-size offline simulation environments within which the agent's policy can be conveniently reinforced to reduce the probability of reiterating safety violating behaviors during subsequent online exploration. As new evidence from online exploration is gathered the abstract representation is incrementally updated and additional offline phases can be enforced when necessary, until an acceptable safety exploration rate is achieved. Overall, our strategy aims at migrating most trial-and-error risks to the offline training phases, while discouraging the repeated exploration of risky behaviours during online learning. As new evidence is collected during online exploration, the abstract representation is incrementally updated and the current \textit{value} the agent expect from each action is used to prioritise the synthesis of more relevant counterexample-guided simulations.

\emph{Our main conceptual contribution in this paper is the use of probabilistic counterexamples to automatically synthesise small-scale simulation submodels where the agent can refine its policy to reduce the risk of violating a safety requirement during learning}.
In particular, we 1) propose a conservative geometric abstraction model representing safety-relevant experience collected by the agent at any time during online exploration, with theoretical convergence and accuracy guarantees, suitable for the representation of both discrete and continuous state spaces and finite action spaces, 2) adapt minimal label set probabilistic counterexample generation~\cite{wimmer2013high} to generate small-scale submodels for the synthesis of offline agent training environments aimed at reducing the likelihood of violating safety requirements during online exploration, and 3) a preliminary evaluation of our method to enhance Q-Learning~\cite{watkins1992q} and DQN~\cite{mnih2013playing} agents on problems from literature and the OpenAI Gym, demonstrating how it achieves comparable cumulative rewards while increasing the exploration safety rate by an average of 40.3\% compared with QL/DQN, and of 29.1\% compared with previous related work~\cite{hasanbeig2018logically,HasanbeigKA22}.
 \section{Background}\label{secPre} 
\subsection{Problem Framework}\label{secFramework}

\begin{definition}{\textbf{Markov Decision Process (MDP).}}
An MDP~\cite{bellman1957markovian} is a tuple $(S, A, s_{0}, P, R, L)$, where $S$ is a set of states, $A$ is a finite set of actions, $s_{0}$ is the initial state, $P: S \times A \times S \rightarrow [0, 1]$ is the probability of transitioning from a state $s \in S$ to $s'\in S$ with action $a \in A$, $R: S \times A \rightarrow \mathbb{R}$ is a reward function and $L: S \rightarrow 2^{\textit{AP}}$ is a labelling function that assigns atomic propositions (\textit{AP}) to each state. 
\end{definition}

\noindent A state in $S$ is typically represented by a vector of finite length $n_S \geq 1$. A state space is \emph{discrete} if the elements of the vector are countable, where we assume $S \subseteq \mathcal{Z}^{n_S})$, continuous if $S \subseteq \mathbb{R}^{n_S}$, or hybrid if some elements are discrete and others are continuous. When possible, we omit the cardinality $n_S$ for readability.

\begin{definition}{\textbf{Trace.}}
A finite trace (also called path or trajectory) through an MDP is a sequence $\sigma = s_0, a_0, s_1,$ $ a_1, \dots s_i, a_i, \dots s_n$, where $s_0$ is the initial state and $P(s_i, a_i, s_{i+1}) > 0$.
\end{definition}

\begin{definition}{\textbf{Policy.}}
    A (deterministic) policy $\pi: S \rightarrow A$ selects in every state $s$ the action $a$ to be taken by the agent. 
\end{definition}

\noindent \textbf{Q-learning} (QL)~\cite{watkins1992q} is a reinforcement learning algorithm where an agent aims at finding an optimal policy $\pi^*$ for an MDP that maximises the expected cumulative reward. Given a learning rate $\alpha \in (0, 1]$ and a discount factor $\gamma \in (0, 1]$, such that rewards received after $n$ transitions are discounted by the factor $\gamma^n$, the agent learns a value function Q based on the following update rule:
\begin{equation}
    Q_{t}(s, a) = (1 - \alpha)Q_{t-1}(s, a) + \alpha(R(s, a) + \gamma \max_{a' \in A} Q_{t-1}(s', a'))
\end{equation}
\noindent The optimal Q-function $Q^{*}$ satisfies the Bellman optimality equation:
\begin{equation}
    Q^{*}(s, a) = \E[R(s, a) + \gamma \max_{a' \in A} Q^{*}(s', a') | s, a]
\end{equation}

\noindent For finite-state and finite-action spaces, QL converges to an optimal policy as long as every state action pair is visited infinitely often~\cite{watkins1992q}, but it is not suitable for learning in continuous state spaces. For continuous state spaces instead, the Deep Q-Learning method~\cite{mnih2013playing} parameterises Q-values with weights $\theta$ as a Q-network and the learning process is adapted to minimising a sequence of loss function $L_{i}$ at each iteration $i$ (cf. Algorithm 1 in~\cite{mnih2013playing}):
\begin{equation}
    L_{i}(\theta_{i}) = \E [(y_{i} - Q(s, a ; \theta_{i}))^2],
\end{equation}
\noindent where $y_{i} = \E[(R(s, a) + \gamma \max_{a' \in A} Q(s', a';\theta_{i-1}) | s, a]$.

During learning, the agent selects the next action among those available in the current state at random with probability $\epsilon_{QL} >0$ while with probability $1 - \epsilon_{QL}$ it will select an action $a$ yielding $Q^*(s,a)$.

\begin{definition}{\textbf{Optimal Policy.}}
    An optimal policy $\pi^{*}$, in the context of Q learning and DQN, is given by $\pi^{*} = \arg \max_{a \in A} Q^{*}(s,a), \forall s \in S$.
\end{definition}

\subsection{Probabilistic Model Checking and Counterexamples}\label{secProbModelChecking}
Probabilistic model checking is an automated verification method that, given a stochastic model -- an MDP in our case -- and a property expressed in a suitable probabilistic temporal logic, can verify whether the model complies with the property or not~\cite{baier2008principles}. 
In this work, we use Probabilistic Computational Temporal Logic (PCTL)~\cite{hansson1994logic} to specify probabilistic requirements for the safety of the agent. The syntax of PCTL is recursively defined as:
\begin{equation}
\Phi := true \mid \alpha \mid \Phi \wedge \Phi \mid \neg \Phi \mid P_{\bowtie p} \varphi
\end{equation}
\begin{equation}
\varphi := X \Phi \mid \Phi U \Phi \end{equation}
A PCTL property is defined by a state formula $\Phi$, whose satisfaction can be determined in each state of the model. $true$ is a tautology satisfied in every state, $\alpha \in AP$ is satisfied in any state whose labels include ($\alpha \in L(s)$), and $\wedge$ and $\neg$ are the Boolean conjunction and negation operators. The modal operator $P_{\bowtie p} \varphi$, with $\bowtie \in \{<, \leq, \geq, >\}$ and $p \in [0,1]$, holds in a state $s$ if the cumulative probability of all the paths originating in $s$ and satisfying the path formula $\varphi$ is $\bowtie p$ \emph{under any possible policy}. The Next operator $X \Phi$ is satisfied by any path originating in $s$ such that the next state satisfies $\Phi$. The Until operator $\Phi_{1} U \Phi_{2}$ is satisfied by any path originating in $s$ such that a state $s^\prime$ satisfying $\Phi_2$ is eventually encountered along the path, and all the states between $s$ and $s^\prime$ (if any) satisfy $\Phi_1$. The formula $true \ U \ \Phi$ is commonly abbreviated as $F \Phi$ and satisfied by any path that eventually reaches a state satisfying $\Phi$. A model $M$ satisfies a PCTL property $\Phi$ if $\Phi$ holds in the initial state $s_0$ of $M$~\cite{baier2008principles}.
 
PCTL allows specifying a variety of safety requirements. For simplicity, in this work, we focus on safety requirements specified as upper-bounds on the probability of eventually reaching a state labelled as \texttt{unsafe}:

\begin{definition}{\textbf{Safety Requirement.}}\label{def:safetyPCTL} Given a threshold $\lambda \in (0,1]$, the safety requirement for a learning agent is formalised by the PCTL property $P_{\leq \lambda} \  [F \ $ $\texttt{unsafe}]$, i.e., the maximum probability of reaching a $s \in S$ such that $\texttt{unsafe}\in L(s)$ must be less than or equal to $\lambda$.
\end{definition}

\begin{definition}{\textbf{Counterexamples in Probabilistic Model Checking.}}
A \\ counterexample is a minimal possible sub-model $M_{cex} = (S_{cex}, A_{cex}, s_{0}, P_{cex})$ derived from the model $M$, where $S_{cex}$, $A_{cex}$ are of subsets $S$ and $A$, containing violating behaviours of a PCTL property from the initial state $s_{0}$ in $M$.
\end{definition}

When a model $M$ does not satisfy a PCTL property, a counterexample can be computed as evidence of the violation~\cite{counterexampleGeneration2009}. In this work, we adapt the \textit{minimal critical label set} counterexample generation method of~\cite{wimmer2013high}.

\noindent The computation of a minimal possible sub-model requires the solution of a mixed-integer linear optimisation problem that selects the smallest number of transitions from the state-action space original model that allows the construction of violations. 
An extensive description of the counterexample generation algorithm, including a heuristic to bias the counterexample generation towards including actions that a tabular Q-learning agent is more likely to select is included in Appendix~\ref{apxCex}.

\noindent\textbf{Generating multiple Counterexamples.} For an MDP violating the safety requirement, there can exist, in general, multiple counterexamples (both with minimal or non-minimal sizes), each potentially highlighting different policies that lead to requirement violations~\cite{vcevska2019counterexample}.

In this work, we use counterexamples to guide the generation of offline training environments where the agent learns to reduce the value of actions that may eventually lead to the violation of safety requirements. We therefore aim at generating multiple, \emph{diverse counterexamples} (if they exist), while keeping each of them at a small size for faster training. Given a counterexample, a different one can be obtained by adding a blocking clause to the minimisation problem, i.e., forcing the optimiser to exclude one or more previously selected action pairs (by imposing the corresponding selector variables $x_\ell=0$ in the optimisation problem of Appendix~\ref{apxCex}). Hence, we can systematically add (an increasing number of) blocking clauses to obtain multiple diverse counterexamples that jointly provide a more comprehensive representation of the different violating behaviors the agent explored at any time. \section{Counterexample-guided Reinforcement Learning}\label{secMethods}
We assume that the agent does not have prior knowledge about the environment. In particular, it will only discover unsafe states upon visiting them during exploration. During the learning process the agent will iteratively interact with either the actual environment (\emph{online} exploration) or with a counterexample-guided \emph{offline} simulation.
The online phases aim at exploring the actual environment and improving the agent's policy expected reward, while acquiring information to build and continuously refine an abstract, compact representation of the control problem. The offline phases expose the agent to simulated, small-size, environments, within which the agent can revise its policy to penalise decisions that may lead to safety violations during the subsequent online phases. 
In the remaining of the section, we first show how to construct and update an abstract finite MDP that compactly represents safety-relevant aspects of the (parts of) environment explored by the agent at any time (sec.~\ref{SecAbs}). 
Then, in sec.~\ref{secCex}, we introduce the main learning algorithm with the online-offline alternation scheme, and discuss the main challenges of the offline learning phases.

\subsection{Safety-relevant State-space Abstraction}\label{SecAbs}
For simplicity, let us assume the agent has no information about the topology of the state space $S$ at the beginning of the exploration. Each online interaction with the environment (\emph{episode}) can be described by the trace of the states visited by the agent and the actions it took. Besides the reward associated with each state-action pair, we assume states in the trace can be assigned a set of labels. Labels represent properties of the state related specific to the learning problem, e.g., a goal has been reached. We assume a special label \texttt{unsafe} labels the occurrence of unsafe situations the agent should aim to avoid. W.l.o.g., we assume an episode terminates when the agent enters an \texttt{unsafe} state. 
We will refer to the states of the online environment as \emph{concrete} states.

In this section, we propose an abstraction procedure to construct a finite, abstract MDP that retains sufficient information about the explored concrete environment to enable the synthesis of abstract counterexamples to the safety requirement. Each counterexample will therefore be an abstract representative of a set of possible safety violating behaviors that can happen in the concrete environment. To maintain the size of the abstract MDP tractable -- especially in the presence of continuous concrete state spaces -- the abstraction will retain only (approximate) safety-relevant information. 

We assume that any state not labeled as \texttt{unsafe} is \texttt{safe} to explore and that the \texttt{unsafe} label is time-invariant. Furthermore, the abstraction must preserve at any time a \textbf{safety invariant}: \textit{every explored unsafe concrete state should be mapped to an unsafe abstract state}. The finite abstract state space must therefore separate safe and unsafe regions of the concrete space, with only safe concrete states possibly misclassified as unsafe but not vice versa. 

Inspired by the idea of casting the learning of geometric concepts as a set cover problem in~\cite{bshouty1998noise,sharma2013verification}, we frame the separation task as a minimal red-blue set cover problem to abstract the explored concrete state space as a finite set of disjoint boxes or polyhedra, each expressed as a set of logical constraints and defined as the intersection of a finite set of hyperplanes.

\noindent To formalise the construction of the abstract state-space, we first introduce the notion of coverage of a concrete state by a polyhedra predicate.

\begin{definition}{\textbf{Coverage of a polyhedra predicate}}\label{def:coverage}
 Let $\bar{S} \subseteq S = \{\vs_{0}, \vs_{1}, \ldots,$ $\ \vs_{n}\}$ be the set of all explored concrete states, a particular state $\vs \in \bar{S}$ is covered by a (polyhedra) predicate $C_{i}$ if $\vs \in C_{i}$, where $C_{i} = \{\vs \ | \ \boldsymbol{\omega} \vs + \vb \leq 0\}$, in which $\boldsymbol{\omega}$ represents the vector of slopes and $\vb$ represents the vector of biases corresponding to half-spaces enclosing the predicate.
\end{definition}

\begin{remark}
 The general affine form of the predicate $C_{i}$ accounts for a variety of common numerical abstract domains, including, e.g., boxes (intervals), octagons, zonotopes, or polyhedra. In this work, we fix $\omega=1$, i.e., restrict to hyper-boxes. We allow the user to specify a minimum size $d > 0$, which ensures that no dimension of the box will be reduced to a length smaller than $d$. This coarsening of the abstract domain struck a convenient trade-off between computational cost and accuracy of the abstraction in our preliminary experiments (see Appendix~\ref{apxTheo} for additional discussion); the restriction can be lifted for applications requiring different trade-offs~\cite{abstractDomains}.
\end{remark}

\noindent The identification of a finite set of predicates that allow separating the concrete state space preserving the safety invariant can thus be reduced to the following:

\noindent\textbf{Minimal Red-Blue Set Cover Problem.} 
Let $\bar{S} \subseteq S  = \{\vs_{0}, \vs_{1}, \ldots , \vs_{n}\}$ be the set of all explored concrete states and $U = \{\vu_{0}, \vu_{1}, \ldots, \vu_{m}\}$ be the set of explored states assigned the \texttt{unsafe} label, find the minimal set $C = \{\cup_{i=1} C_{i}\}$ s.t. every element $\vu \in U$ is covered by some predicate $C_{i}$, with an overall false positive rate $\text{fpr} \leq f \in (0,1]$ for safe concrete states ($ \vs \in \bar{S} \setminus U$) covered by $C$.

\begin{remark}
In general, $f$ cannot be zero, since the concrete state space, whether discrete or continuous, may not be perfectly partitioned by a finite set of polyhedra predicates, with smaller values of $f$ possibly resulting in a larger number of predicates $|C|$. 
\end{remark}

\vspace{1mm}
\noindent To solve this optimisation problem, we employ a branch and bound method~\cite{lawler1966branch} to systematically enumerate possible combinations of predicates. The solution set guarantees all the unsafe concrete states are covered by a Boolean combination of predicates in $C$, while safe concrete states may also be covered by some predicate $C_i$, with a prescribed maximum tolerable rate $f$.

\begin{definition}{\textbf{Safety-relevant Abstraction MDP.}}
    A safety-relevant abstraction MDP $M_a$ is a tuple $(S_a, A_a, s_{a0}, R_a, P_a, L_a)$, where $S_a$ is the abstract state space, which is the partition of the concrete state space $S$ induced by the boundaries of $C_{i} \in C$ from the solution of the minimal set cover above, $A_a$ is the set of applicable actions, $s_{a0}$ is the initial abstract state, $P_a: S_a \times A_a \times S_a \to [0,1]$ is a probability transition function, $R_a$ is the abstract reward function (which will be defined later), and $L_a: S \to \{\texttt{safe}, \texttt{unsafe}\}$ is a labelling function.
\end{definition}

\noindent $M_a$ is constructed from the concrete traces collected during online learning, with the satisfaction of the predicates $C_i$ determining the abstraction of concrete states and the abstract transition function is estimated accordingly from the frequencies observed in the concrete traces. The abstraction must preserve the \emph{safety invariant}, therefore it may overapproximate explored unsafe regions, but not underapproximate them. Initially, the entire state space is assumed safe to explore. As new traces are collected during online exploration, the abstract model is incrementally updated: when a concrete state is found to be unsafe, the hyperbox containing its numerical vector representation is split to correct the classification (after the concrete state is wrapped around with a hyperbox of minimal size $d$, if $d>0$).

The incremental branch-and-bound refinement of the abstraction could lead to excessive fragmentation of the abstract state space, making it intractably large, particularly for the purpose of counterexample generation. To mitigate this issue, we merge adjacent states, i.e., abstract states sharing at least one separating hyperplane, into a single abstract state, adapting the general notion of probabilistic approximate $\epsilon-$simulation~\cite{desharnais2008approximate} as in the following definition:
\begin{definition}\label{defEpsilonSim}{\textbf{Adjacent $\epsilon$-simulation}:}
Let $S_l$ be the partitions of $S_a$ induced by the equivalence relation $s \sim s' \text{ iff } L_a(s)=L_a(s')$. Then, for a given $\epsilon \in [0,1]$, two adjacent states $s \in S_{a}$ and $s' \in S_{a}$ are $\epsilon$-similar if $(\exists s_l \in S_l) (s \in s_l, s' \in s_l)$ and $(\forall s_l \in S_l) (\left|P_{a}(s, a, s_{l}) - P_{a}(s', a, s_{l})\right| \leq \epsilon, \forall a \in A_{a})$.     
\end{definition}
The $\epsilon$-simulation in def.~\ref{defEpsilonSim} induces a hierarchical merging scheme. Let level $l_0$ contain the initial abstract states, which partition the explored concrete state space into a finite set of boxes -- one per abstract state -- which are labeled as either safe or unsafe. $\epsilon$-similar adjacent states from level $l_i$ are merged into a single state at level $l_{i+1}$, until no further merge is possible. Besides reducing the number of abstract states, and in turn the cost of generating counterexamples, this hierarchical merging scheme brings the indirect benefit of more aggressively merging abstract states corresponding to safe regions of the concrete state space, while preserving a finer-grained approximation of concrete state space regions in proximity of explored unsafe states, as discussed in Appendix~\ref{apxEval}.

\noindent\textbf{Counterexample-guided Simulation.} If a probabilistic safety requirement can be violated, one or more counterexamples $M_{cex}$ can be generated from the abstract model $M_a$, where each counterexample includes a (near-)minimal subset of the abstract state-action space. We then use each counterexample as a guide to build an offline, simulation environment where the agent can update its Q-values towards avoiding eventually reaching an unsafe state.

Starting from the initial concrete state $s_0$, the abstract state $s_{cex}$ in $M_{cex}$ corresponding to the current concrete state is computed. By construction, each counterexample selects one action $a$ from state $s_{cex}$. The abstract transition is randomly simulated according to the transition function $P_{cex}$ from $(s_{cex}, a)$ and an abstract destination state $s_{cex}^\prime$ is identified. Such abstract state is concretised by sampling from the past concrete traces a transition $(s, a, s^\prime)$ where $s \in s_{cex}$ and $s^\prime \in s^\prime_{cex}$. If $s^\prime_{cex}$ is an unsafe state, a penalty (negative reward; the impact of its magnitude is further discussed in Appendix tab.~\ref{tblHyperparameters}) is given to the agent, which has the transitive effect of re-weighting also the Q-value of the actions that led the agent to the current state. The simulation traces can be used by both Q-learning and DQN agents. The simulation terminates when an unsafe state is reached (penalty) or when we fail to concretise an abstract transition, which may happen when concrete safe states are misclassified as unsafe in the abstraction, but there is no actual transition to unsafe states from them. In the latter case, the simulation trace is discarded (no reward). While every simulation within the counterexample is designed to eventually reach an unsafe state with probability 1 by construction~\cite{wimmer2013high}, to avoid excessive length of a simulation, it can be practical to set an arbitrary, large bound on the maximum number of steps per run as additional termination criterion.

Multiple counterexamples, and corresponding simulations, can be generated up to a maximum simulation budget allowed by the user, adding blocking clauses in random order as described previously. Each counterexample is typically of small-size, which results in short simulation traces, thus reducing the overall cost of each offline learning experience.

\subsection{Online-Offline Learning with Counterexample Guidance}\label{secCex}
Algorithm~\ref{alg:alg1} summarises the main steps of our online-offline learning method with counterexample guidance. We initially assume no knowledge about the environment is given to the agent: both the abstract model $M_a$ and the set of explored paths $D$ are empty (line~\ref{algInit}). If prior knowledge was available, either in the form of an initial abstraction or of previously explored paths, $M_a$ and $D$ can be initialised accordingly.

\noindent\textbf{Online learning.} The \texttt{onlineQLearning} procedure (line~\ref{algOnlineLearning}) lets the agent operate in the concrete environment with either tabular Q-Learning in discrete state space or DQN in continuous state space. We augment the exploration with a sequential Bayesian hypothesis testing (line~\ref{algBayesianHypothesis}) that monitors the frequency of violations of the safety requirement, by incrementally updating after each online episode a Beta distribution that estimates the probability of violation~\cite{downey2021think}. If the odds of such probability exceeding $\lambda$ is larger than a prescribed Bayes factor $\beta$, the online learning phase is interrupted. The updated Q-values/Q-network of the agent are stored and the set of explored traces $D$ is updated (line~\ref{lnUpdateUnsafe}).

\noindent\textbf{Offline learning.} If the online learning phases has been interrupted because by the Bayesian test (line~\ref{algIfUnsafe}), an offline learning phase is triggered to reinforce the avoidance of discovered unsafe behaviors in future online exploration.

First, the abstraction $M_a$ is updated with the current set of online traces $D$ (line~\ref{algUpdateAbstraction}) to support the generation of current counterexamples. While there are theoretically a finite number of counterexample submodels~\cite{wimmer2013high}, for large $M_a$ it could be computationally too expensive; instead, up to a maximum number $N_{cex}$ of counterexamples is generated at each iteration. The addition of random blocking clauses (as described in sec.~\ref{secProbModelChecking}) will increase the diversity of the counterexamples within and across different offline learning phases.

The offline simulation traces synthesised from each $M_{cex}$ (line~\ref{algSimulationTraces}) as described in the previous section are used by the agent to update its Q-values/Q-network (line~\ref{algOfflineUpdate}), thus penalising the selection of eventually unsafe actions before the next online learning phase begins. Notice that the Bayesian hypothesis test is re-initialised before the next online learning phase (line~\ref{algBayesianHypothesis}) since the agent is expected to behave differently after an offline learning phase. 

\noindent\textbf{Discussion.} 
The interleaving of offline and online learning phases aims at reducing the frequency of unsafe events during the exploration of the environment. This goal is pursued by synthesising simulation environments from counterexamples to the safety requirement computed from an abstraction of the state space explored online. Notably, the offline phases never preclude the exploration of an action during the following online phases, rather they reduce the likelihood of selecting actions that may eventually lead to reaching unsafe states by lowering their Q-values. Due to space limitations, we report here the two main results related to the convergence of our abstraction method and of the online-offline learning process, and defer a more extensive discussion to Appendix~\ref{apxTheo}.

Counterexample guidance relies on the abstraction constructed from online exploration phases, which classifies every region of the explored state space as either safe or unsafe. While by construction the abstraction preserves the safety invariant (every explored concrete unsafe state is mapped to an abstract unsafe state), the quality of the offline guidance relies also on controlling the misclassification error of safe concrete regions, which may unduly penalise the exploration of safe states.

\begin{restatable}{proposition}{propthree}\label{prop3}
The maximum misclassification error of a concrete safe state into an abstract unsafe state can eventually be reduced below an arbitrary bound $0 < \bar{u} \leq 1$ with probability at least $1-\delta$ ($0 < \delta <1$) throughout the exploration.
\end{restatable}

\noindent Further empirical analysis of the convergence of the abstraction and the impact of the abstraction parameters is provided in Appendix~\ref{apxEval}.

Finally, the following proposition states that the introduction of counterexample guidance does not preclude the convergence of the overall learning process to a maximal reward policy that satisfies the safety requirement, if such policy exists.

\begin{restatable}{proposition}{propfour}\label{prop4}
If there exist maximal-reward policies that satisfy the safety requirement, then the online-offline learning process eventually converges to one of them.
\end{restatable}

\noindent Further discussion of the convergence properties of the offline-online learning process are included in Appendix~\ref{apxTheo}, including elaborating on the validity of the two propositions above. 
In the next section, we instead report on our preliminary experimental evaluation of the performance of our counterexample-guided learning process.

\SetKwComment{Comment}{//}{}

\newcommand\mycommfont[1]{\footnotesize\ttfamily#1}
\SetCommentSty{mycommfont}

\begin{algorithm}[!htb]
\caption{CEX-Guided Reinforcement Learning}\label{alg:alg1}
\textbf{Input:} Safety requirement bound $\lambda > 0$, max number of CEX $N_{cex}$, num simulations per CEX $NS_{cex}$, Bayes factor $\beta$, penalty for offline learning traces $penalty$\\
\textbf{Output:} Synthesised Policy $\pi$\\
Initialise $M_a, D \gets \emptyset$\label{algInit}\;
Initialise $Q$ to random values\;
\While{not converged}{

$\delta_{bht} \gets \texttt{BayesianHypothesisTester}(\lambda, \beta)$\label{algBayesianHypothesis}

Q, $D_{online} \gets \texttt{onlineQLearning}(Q, \delta_{bht})$\label{algOnlineLearning} \;

$D \gets D \cup D_{online}$\label{lnUpdateUnsafe}\;

\If{$\delta_{bht} > \beta$\label{algIfUnsafe}}{
    \Comment{cex-guided offline learning}
$M_a \gets \texttt{refine}(M_a, D)$\label{algUpdateAbstraction}\tcp*{update the abstract MDP}

\For{up to $N_{cex}$\label{algOfflineLearning}}{
 
 $M_{\texttt{cex}} \gets \texttt{cexGeneration}(M_a, \lambda)$

 simTraces $\gets$ collectSimulationTraces($M_{cex}, Q, NS_{cex})$\;\label{algSimulationTraces}
 Q $\gets$ Q-value/Q-network update(simTraces, $penalty$)\;\label{algOfflineUpdate}
}
}
}
$\pi \gets \arg \max_{a}Q$\;
\end{algorithm} \section{Evaluation}\label{secEvaluation}
In this section, we present a preliminary experimental evaluation of the performance of our method from two perspectives: 1) the improvement in the exploration safety rate, and 2) the impact on the cumulative reward achieved by the agent. Finally, we briefly discuss the overhead of counterexample guidance and make some observations on the policies it synthesises. Additional experimental results and discussion, including on abstraction effectiveness and sensitivity to hyperparameters can be found in Appendix~\ref{apxEval}.

\noindent\textbf{Environments. }
We consider four environments: \texttt{DiscreteGrid} from the implementation of~\cite{HasanbeigKA22}, the slippery \texttt{FrozenLake8x8} from OpenAI Gym~\cite{brockman2016openai}, \texttt{HybridGrid} -- where we change the state space from discrete to continuous with the same layout in~\cite{HasanbeigKA22}, and \texttt{MarsRover}~\cite{hasanbeig2020deep} (in particular, the exploration of the melas chasma in the Coprates quadrangle~\cite{mcewen2014recurring}). In all the environments, the agent decides a direction of move between \texttt{up}, \texttt{down}, \texttt{left}, and \texttt{right}. We define the objective of the agent as finding a policy with maximum Q-value while avoiding unsafe behaviours with intolerably high probability $\lambda$ during exploration. Specifically, in the \texttt{FrozenLake8x8}, the agent aims to find a walkable path in a 8x8 grid environment with slippery actions while avoiding entering states labelled with \textit{H}. With the slippery setting, the agent will move in the intended direction with a probability of only 1/3 else will move in either perpendicular directions with equal probabilities of 1/3 respectively. In the \texttt{DiscreteGrid} and \texttt{HybridGrid}, the agent aims to reach the states labelled with \texttt{goal1} and then states labelled with \texttt{goal2} in a fixed order while avoiding entering any states labelled with \texttt{unsafe} along the path, with $15\%$ probability of moving to random directions at every action. In \texttt{HybridGrid}, the distance covered in a move is also randomly sampled from a Gaussian $\mathcal{N}(2,0.5)$, thus choosing the same direction from a state may reach different states. In the \texttt{MarsRover} environment, the agent aims to find one of the target states labelled with \texttt{goal} while avoiding reaching the unsafe regions, which in this scenario cannot be perfectly abstracted by boxes or other affine abstract domains. Following~\cite{lcrl_tool}, the distance covered in each move is sampled uniformly in the range $(0, 10)$, with the addition of further uniform noise from $\mathcal{U}(-0.1, 0.5)$.

\noindent\textbf{Baselines.} 
We compare the learning performance of our method with classical Q-Learning and DQN~\cite{mnih2013playing} as the baseline for discrete and continuous MDPs, respectively. For discrete MDPs, we further compare our method with~\cite{hasanbeig2018logically} (referred to as QL-LCRL in the following), using the same set of hyper-parameters as provided in their implementation and the associated tool paper~\cite{lcrl_tool,HasanbeigKA22}. Given an automata corresponding to an LTL property, QL-LCRL guides the agent's exploration of an initially unknown MDP by reshaping on-the-fly the reward function to encourage the exploration of behaviors that satisfy such property. In this application QL-LCRL will encourage a safe exploration by discouraging reaching \texttt{unsafe} states.

\noindent\textbf{Implementation and parameters.} We implemented a standard tabular Q-learning in Python and used the DQN implementation from OpenAI Gym Baselines~\cite{openaibaselinesdqn}. We parameterise the abstraction process with a learning rate $\alpha$ and a discount factor $\gamma$ for the Q-value/Q-Network updates, and $\epsilon$ for adjacent $\epsilon$-bisimulation. The agents move within Cartesian planes of sizes $|S|$ and the minimisation of the abstract models reduces the state space to $|S_{a}|$, where the minimum size of each box is set to 1 and 0.01 for the discrete and the continuous environments, respectively. We set the safety specification parameter $\lambda$ according to the intrinsic uncertainty in the respective environments. A summary of the parameters used for each environment is reported in the left side of tab.~\ref{tab:learning_results} (additional parameters are discussed in Appendix tab.~\ref{tab:hyperparamters} to ease reproducibility). We require at least 50 samples to be collected by the Bayesian hypothesis test before it can trigger offline training to reduce false positive triggers.

\begin{table}[h]
\centering
\begin{tabular}{|l||c|c|c|c|c|c|c||c|c|c|c|c|c|}\hline
\tiny{Environment} &\makebox[1em]{\tiny{$|S|$}}&\makebox[1.5em]{\tiny{$|S_{A}|$}}&\makebox[1em]{\tiny{$\lambda$}}&\makebox[1em]{\tiny{$\alpha$}}&\makebox[1em]{\tiny{$\gamma$}}&\makebox[1em]{\tiny{$\epsilon$}}&\makebox[1em]{\tiny{fpr}}&\makebox[3em]{\tiny{\makecell{QL/DQN \\ safe.prob}}}&\makebox[3em]{\tiny{\makecell{QL/DQN \\ reward}}}&\makebox[3em]{\tiny{\makecell{QL-LCRL \\ safe. prob}}}&\makebox[3em]{\tiny{\makecell{QL-LCRL \\ reward}}}&\makebox[3em]{\tiny{\makecell{Q-CEX\\safe. prob}}}&\makebox[3em]{\tiny{\makecell{Q-CEX \\ reward}}}\\\hline\hline
\tiny{DiscreteGrid} &\tiny{3200}& \tiny{30} & \tiny{0.2}& \tiny{0.9}&\tiny{0.9}&\tiny{0.01}&\tiny{0.05}&\tiny{0.299}&\tiny{0.924}&\tiny{0.387}&\tiny{0.99}&\tiny{0.572}&\tiny{0.99}\\\hline
\tiny{FrozenLake8x8} &\tiny{64}& \tiny{50}& \tiny{0.35} & \tiny{0.1}&\tiny{0.9}&\tiny{0.01}&\tiny{0.05}&\tiny{0.339}&\tiny{0.55}&\tiny{0.401}&\tiny{0.54}&\tiny{0.443}&\tiny{0.59}\\\hline
\tiny{HybridGrid} &\tiny{$\infty$}& \tiny{41} & \tiny{0.2}& \tiny{5e-4}&\tiny{0.9}&\tiny{0.01}&\tiny{0.05}&\tiny{0.548}&\tiny{0.753}&-&-&\tiny{0.687}&\tiny{0.78}\\\hline
\tiny{MarsRover} &\tiny{$\infty$}& \tiny{149} & \tiny{0.2}& \tiny{5e-4}&\tiny{0.9}&\tiny{0.05}&\tiny{0.15}&\tiny{0.681}&\tiny{0.947}&-&-&\tiny{0.775}&\tiny{0.86}\\\hline
\end{tabular}
\caption{Learning results with corresponding hyperparamters in CEX-guided Learning}\label{tab:learning_results}
\end{table}

\begin{figure}[!h]
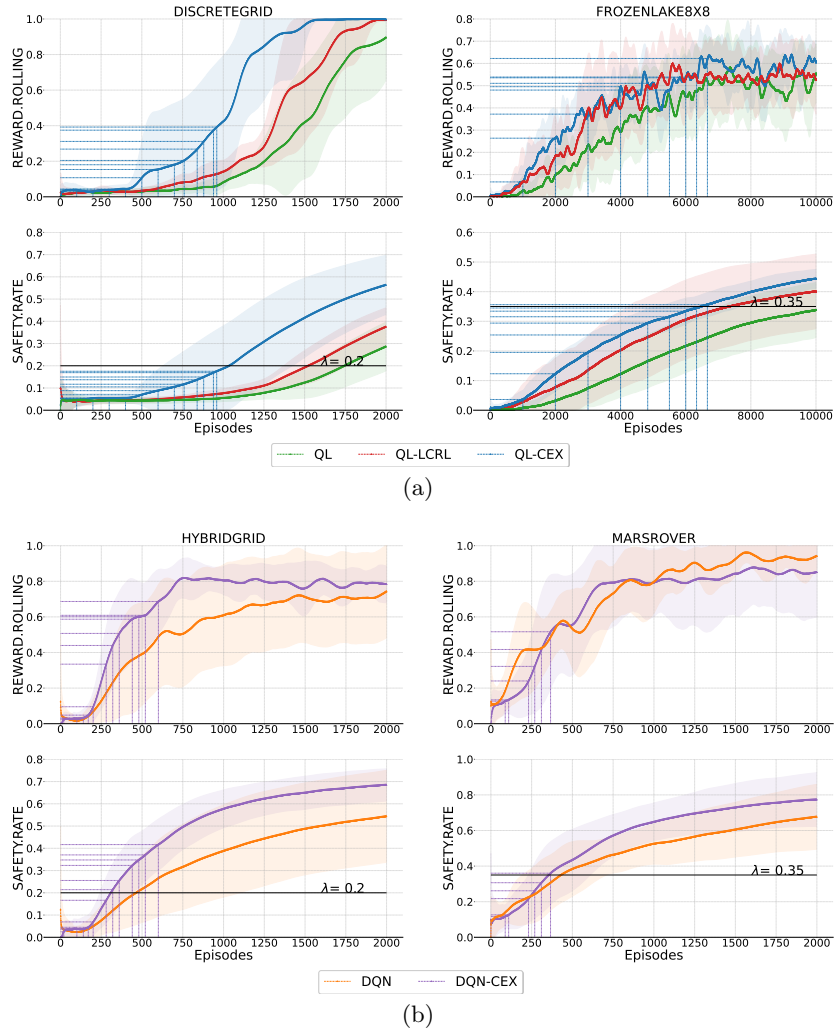
 
\centering
\subfigure[]{\includegraphics[width=0.9\columnwidth]{imgs/training_ql.pdf}}\quad
\subfigure[]{\includegraphics[width=0.9\columnwidth]{imgs/training_dqn.pdf}}

\caption{Average over 10 runs of the accumulated safety rates and rolling rewards in QL/DQN, QL-LCRL~\cite{HasanbeigKA22} and QL/DQN-CEX (our method). The shaded region shows $\pm$ standard deviation.}\label{fig:training_results}
\end{figure}

\noindent\textbf{Experimental Results.} 
Fig.~\ref{fig:training_results} shows the accumulated safety rates (bottom) and the rolling average of accumulated rewards, indicating the real-time learning performance of the agent. The line is the average across 10 runs, while the shaded region around it is the standard deviation. We do not provide any prior information to the agent. 
The dashed vertical lines in the figure indicate the average episode number where an offline learning phase in QL/DQN-CEX (our method) is triggered, and the solid horizontal line indicates the target safety rate in corresponding safety specifications. The cumulative rewards of different methods converge to similar values, demonstrating that the performance under guidance (QL-LCRL and Q-CEX) achieves cumulative rewards comparable to the baseline QL and DQN method. 
As expected, providing additional guidance to discourage actions that may lead to reaching unsafe states with QL/DQN-CEX or QL-LCRL improves the safety rate of the exploration, with QL/DQN-CEX achieving on average higher safety rates.

In \texttt{DiscreteGrid}, the safety rate of online learning in our method exceeds the threshold much faster than other methods. This is due to the rapid convergence of the abstraction thanks to the grid layout that can be accurately and efficiently abstracted (and minimised). In turn, no further offline phases were required after the first 1000 episodes.

In \texttt{FrozenLake8x8}, more online exploration is required to support comprehensive counterexample guidance, partly due to the high uncertainty in the outcome of the actions. Another phenomenon due to high uncertainty is that the agent takes a longer time to reach a stable performance, therefore the offline learning phase is triggered more frequently compared with other environments and also more episodes were required to stably satisfy the safety requirement. 

In \texttt{HybridGrid} and \texttt{MarsRover}, while the baseline eventually achieves a marginally higher cumulative reward in MarsRover, DQN-CEX achieves a higher safety rate -- the number of failures experienced by the agent is inversely proportional to the integral of the safety rate curve, which results in significantly fewer failure events. The frequency of offline training phases also decreases over time. This is not surprising due to a safer exploration being possibly less speculative and slower in exploring the optimal policy. QL-LCRL is not applicable to \texttt{HybridGrid} and \texttt{MarsRover}. Although LCRL~\cite{HasanbeigKA22} applies NFQ-based method for continuous MDPs, NFQ-LCRL trains the agent in a completely offline manner based on randomly sampled data, thus it is not suitable for comparison with our method from the perspective of safe exploration.

\noindent\textbf{Overhead.} Counterexamples generation requires solving a MILP problem on the abstract state space. On a Macbook Air with M1 CPU and 8Gb of memory, the average$\pm$stdev time to solve the optimization problems were 0.71$\pm$0.98s, 0.08$\pm$0.06s, 2.31$\pm$3.94s, and 3.87$\pm$4.07s for \texttt{FrozenLake8x8}, \texttt{DiscreteGrid}, 

\noindent\texttt{HybridGrid}, and \texttt{MarsRover}, respectively. We used Gurobi Optimiser v9.1.0~\cite{gurobi} off-the-shelf. The numbers of counterexamples generated for each offline learning phase were, on average: 11, 6, 19, and 21, respectively. Notice that both counterexample generation and the simulations can be parallelised. While the cost of solving the MILP may be higher, we notice that offline learning is triggered only when the recent online exploration resulted in an unacceptable failure rate. As shown in fig.~\ref{fig:training_results}, thanks to counterexample-guidance, QL/DQN-CEX can achieve the required safety exploration rate much faster, which helps amortizing the initial MILP solution cost. Finally, the optimisation problem might be relaxed to sacrifice the optimality of the solution (i.e., size of the counterexamples) for computation time.
 \section{Related Work}\label{secRelated}
\noindent\textbf{Safe Exploration.}  
\cite{garcia2015comprehensive,kim2020safe,brunke2022safe} provide surveys and taxonomies of recent safe reinforcement learning methods with different emphases. Most existing safe exploration methods assume prior knowledge about the environment or the agent's dynamics ~\cite{moldovan2012safe,dalal2018safe,FultonPlatzer2018}, known safety constraints~\cite{mason2017assured,jansen2018shielded,hasanbeig2018logically,pham2018optlayer}, or utilise expert guidance~\cite{garcia2012safe,zhou2018safety,huang2018learning,prakash2019improving} to provide guarantees of safety constraints satisfaction during exploration. A different class of methods~\cite{sui2015safe,wachi2018safe,liu2020robust} utilises surrogate Gaussian process models to characterise the unknown dynamics and optimise the unknown function with a \emph{prior} model structure. There are only a few methods~\cite{hasanbeig2020deep,bharadhwaj2020conservative} tackling safe exploration without known model structure. \cite{hasanbeig2020deep} focuses on solving continuous, unknown MDPs into sub-tasks using an online RL framework under LTL specification with less emphasis on safety rate but blocking actions unsafe according to the specification, and~\cite{bharadhwaj2020conservative} trains a safety layer/critic used for filtering out probabilistic unsafe actions with an offline dataset. While motivated by the same idea of safer exploration without prior knowledge, our method can be initialised with none or any amount of previously explored paths, meanwhile it converged to cumulative rewards comparably or better than baseline methods.

\noindent\textbf{Offline RL.} 
Offline RL can be seen as a data-driven formulation of RL, where the agent collects transitions using the behaviour policy instead of interacting with the environment~\cite{levine2020offline}. The biggest challenge in offline RL is the bootstrapping error: the Q-value is evaluated with little or no prior knowledge and propagated through Bellman equation~\cite{kumar2019stabilizing}. \cite{wu2019behavior,siegel2020keep} regularise the behaviour policy while optimising to address this issue. ~\cite{buckman2020importance,kumar2020conservative} alternatively update the Q values in more conservative ways to learn a lower bound of the Q function using uncertainty estimated with the sampled information. From the safe RL perspective, \cite{urpi2021risk} optimises a risk-averse criterion using data previously collected by a \textit{safe} policy offline, and~\cite{xu2022constraints} learns a constrained policy maximizing the long-term reward based on offline data without interaction in the concrete environment using a constrained penalised Q-Learning method. These methods have similar motivation of combining offline learning with risk-averse RL as ours, while focusing on continuous control setting instead. Besides utilising offline learning to reduce unsafe online exploration, we alternate online and offline learning to keep the abstract knowledge about the environment up to date and increase the risk aversion of the agent based on the most current evidence it collected online. \section{Conclusion}\label{secConclusion}
We presented our investigation of a safer model-free reinforcement learning method using counterexample-guided offline training. 
We proposed an abstraction strategy to represent the knowledge acquired during online exploration in a succinct, finite MDP model that can consistently and accurately describe safety-relevant dynamics of the explored environment. Counterexample generation methods from probabilistic model checking are then adapted to synthesise small-scale simulation environments capturing scenarios in which the decisions of the agent may lead to the violation of safety requirements. The agent can then train offline within this minimal submodels by replaying concrete transitions recorded during past online exploration consistent with the counterexample, using a reward scheme focused on reducing the likelihood of selecting actions that may eventually lead to visiting again explored unsafe concrete states. The q-values penalized during the offline training phases implicitly reduce the risk of repeating unsafe behaviors during subsequent online exploration, while newly explored paths feedback information to the next offline learning phase. 

The alternation of online exploration -- and abstraction refinement -- and counterexample guided learning can ultimately lead to higher safety rates during exploration, without significant reduction in the achieved cumulative reward, as demonstrated in our preliminary evaluation on problems from previous literature and the OpenAI Gym. While this paper focused on improving Q-Learning (and the related DQN algorithm), the fundamental framework is not specific to Q-Learning, and we plan to explore its impact on other learning algorithms in future work. 

\vspace{1mm}
\noindent\textbf{Data availability.} An artifact including the prototype Python implementation used for the experiments has been accepted by QEST 2023 artifact evaluation. The implementation of our method is available at Github:~\url{https://github.com/xtji/CEX-guided-RL}.

\clearpage
\bibliographystyle{splncs04}

\newpage
\appendix
\newcommand\qedsymbol{$\blacksquare$}

\section{Appendix}
\subsection{Minimal Counterexamples Generation}\label{apxCex}
\begin{definition}{\textbf{Counterexample of the Safety Specification:}}\label{counterexampleMILP} The solution of the following optimisation problem (adapted from~\cite{wimmer2013high}) is a minimal counterexample of the safety specification $P_{\leq \lambda} \  [F \ \texttt{unsafe}]$:
$${\rm minimise} \ -\frac{1}{2} \omega_0 p_{s_{0}} + \sum_{\ell \in \rm L}{\omega(\ell)x_{\ell}}, \ {\rm, such \ that}$$\label{eqCex}
\begin{align}
&p_{s_{0}} > \lambda \label{eq:caf01} \\
&\forall s \in T. \ \  p_{s} = 1 \label{eq:caf02} \\
&\forall s \in S \setminus T. \ \ \sum_{a \in P(s)} \ \pi_{s, a} \leq 1 \label{eq:constraint5} \\
&\forall s \in S \setminus T. \ p_{s} \leq \sum_{a \in P(s)} \ \pi_{s, a} \label{eq:caf03}\\
&\forall s \in S \setminus T, \ \forall a \in A, \ \forall \ell \in L(s, a, s'). \ \ p_{s, a, s'} \leq x_{\ell} \label{eq:caf04}\\
&\forall s \in S \setminus T, \ \forall a \in A, \ \ p_{s, a,  s'} \leq P(s,a,s') \cdot p_{s'} \label{eq:caf05}\\
&\forall s \in S \setminus T, \ \forall a \in A. \ \ p_{s} \leq (1 - \pi_{s, a}) \ + \sum_{s' : P(s,a,s')>0} p_{s, a, s'}\label{eq:constraint10}\\
&\forall (s, a) \in P_{T}^{Prob} . \ \ \pi_{s, a} = \sum_{\ell \in L(s, a, s')} x_{\ell}\label{eq:constraint11}\\
&\forall (s, a) \in P_{T}^{Prob} \ \ \forall \ell \in L(s, a, s'). \ \ r_{s} < r_{s'} + (1 - x_{\ell})\label{eq:constraint12} \\
\end{align}
\noindent where $S$ is the state space of the model, $T$ is the set of states labelled as \texttt{unsafe}, $p_{s}$ represents the probability of reaching any state in $T$ from state $s$, $\pi_{s,a} \in \{0,1\}$ indicates that action $a$ is selected in state $s$, $p_{s,a,s'}$ represents the probability contribution of the transitions from $s$ to $s'$ via action $a$, where this transition is selected to be part of the counterexample, $\ell \in L(s,a,s')$ is the label identifying a transition from $s$ to $s'$ via action $a$ (not to be confused with the function $L$ labeling states of the model) such that $x_l=1$ iff the transition is included in the counterexample, $x_l=0$ otherwise, and $P_{T}^{Prob}$ represents the set of problematic state-action pairs if the minimal probability of reaching the target state is zero and the maximal probability of reaching the target state is non-zero from these state-action pairs.
\end{definition}

Intuitively, the optimisation problem aims to find a policy $\pi$ that will make the agent violate the safety specification within the smallest counterexample sub-model, composed of all the transitions $(s,a,s')$ from the original model whose corresponding $x_l$ is 1. 

To this goal, eq.~\ref{eq:caf01} requires the probability of reaching an unsafe state from the initial state $s_0$ to be $>\lambda$ (violation of the safety specification); eq.~\ref{eq:caf02} fixes $p_s=1$ for all the unsafe states; eq.~\ref{eq:constraint5} impose the agent to select at most one action $a$ for each state $s$; if no action is chosen in a state $s$, eq.~\ref{eq:caf03} ensures that $p_s=0$. Other constraints for ensuring the minimal size and prevent deadlock loops are also defined~\cite{wimmer2013high}. We further specialise the obejctive based on the approach in~\cite{wimmer2013high} by setting the weights $\omega(\ell)$ of the selector variables $x_\ell$ as one minus the normalised Q-values corresponding to the state-action pair in $\ell$, while $\omega_0 > \max \{\omega(\ell) \mid \forall \ell \in L_{c} \wedge \omega(\ell) > 0 \}$. With this weighting, we encourage the selection of labels with larger Q-Value, i.e., corresponding to the violating behaviours most likely to be selected by the agent, while at the same time minimising the size of the sub-model. Eq.~\ref{eq:caf04} ensures that the contribution of the transition $(s,a,s')$ is 0 if such transition is not included in the counterexample, otherwise, eq.~\ref{eq:caf05}, $p_{s,a,s'}$ is bounded by the probability of the corresponding transitions in the model -- $P(s,a,s')$) times the probability of reaching the target from $s'$ $p_{s'}$. Eq.~\ref{eq:constraint10} ensures that if action $a$ is selected in state $s$, the probability of reaching $T$ from $s$ is bounded by the sum of the probabilities of reaching $T$ from its successors given $a$. Finally, two additional constraints are defined in~\cite{wimmer2013high} to prevent the agent from getting stuck in an infinite loop that would prevent it from reaching $T$. 

\vspace{1mm}
Compared to the general solution in~\cite{wimmer2013high}, for the case of tabular Q-Learning, we heuristically specialise the objective function by assigning to the selection of a state-action pair $(s,a)$ a cost proportional to $-\bar{Q}(s,a)$, where $\bar{Q}(s,a)$ is the normalised where this average Q-value of $a$ over the concrete states represented by an abstract state. Because the minimal size counterexample is, in general, not unique, this additional cost prioritises the inclusion of actions with larger Q-Value, i.e., actions most likely to be selected by the agent, while at the same time minimising the size of the sub-model.

\subsection{On the Convergence of the Abstraction and Learning Processes} \label{apxTheo}

In this section we will discuss the main convergence aspects of the proposed abstraction method and of the alternating online-offline learning process.

\noindent\textbf{Convergence of the abstraction.} 
The core of the proposed abstraction of safety relevant aspects of the explored concrete state space revolves around the solution of a minimal red-blue set coverage by means of polyhedra (restricted to hyperboxes in this work) predicates introduced in sec.~\ref{SecAbs}. We aim to discuss how an abstraction up to an arbitrary accuracy will almost surely eventually be constructed. We remind that the set cover problem allows only the overapproximation of unsafe regions, with unsafe concrete points always mapped to unsafe abstract states, while safe concrete points may possibly be misclassified as unsafe with a maximum prescribed false positive rate.

For simplicity, let us focus the discussion on learning for MDPs with discrete state space. Because we allow hyperboxes of minimum size $d>0$, a continuous space is implicitly discretised, with every sample from a continuous space included within a box of size $d$ or larger.

\begin{restatable}{proposition}{prop1}\label{prop1}
Every reachable concrete state will eventually be reached with probability 1 during online exploration, in particular every reachable unsafe state will eventually be reached.
\end{restatable}

This follows from the fact that the online exploration of the environment is never strictly limited by the use of counterexample guidance. Rather, offline phases aim at reducing the relative Q-value of actions that may eventually lead to the violation of the safety requirement, thus reducing the likelihood of their selection during online learning. Because the agent is always allowed, with a controllable probability, to explore any action from the current state, every reachable state maintains a strictly positive probability of being reached.

\begin{restatable}{proposition}{prop2}\label{prop2}
Every reachable unsafe state will eventually be mapped to an unsafe abstract state.
\end{restatable}

Proposition~\ref{prop2} follows from Proposition~\ref{prop1} and the preservation of the safety invariant, which ensures unsafe concrete states are always mapped to unsafe abstract states. Finally, restated from sec.~\ref{secCex}:

\propthree*

Proposition~\ref{prop3} relies on a PAC learnability argument. At any time during the exploration, the maximal explored state is bounded by the most extreme states that have been explored.

During exploration, the agent can select randomly an action among those available in the current state with probability $\epsilon_{QL} > 0$. This exploration probability may change over time, but should always be strictly larger than zero to ensure every state-action pair can be selected infinitely often to ensure the convergence of Q-Learning (cf. sec.~\ref{secFramework}). Let $\underline{\epsilon_{QL}} > 0$ be the lowerbound of the values  $\epsilon_{QL}$ can take during exploration. Then, given $\underline{\epsilon_{QL}}$ and the concrete MDP's transition relation, the probability of visiting any (reachable) concrete state is bounded from below by a value $\underline{p_{QL}}$. 

Recalling Theorem 2.1 in~\cite{blumer1989learnability}, if a learning concept $L$ has a finite Vapnik–Chervonenkis (VC) dimension, and if $L$ is consistent with a uniform sample size $\max(\frac{4}{\bar{u}}\log\frac{2}{\delta}, \frac{8VC}{\bar{u}}\log\frac{13}{\bar{u}})$, then the prediction error of the learning concept $L$ can be limited to $u \in (0, 1]$, with a probability of at least $1-\delta$.
The expressiveness of $L$ and also VC dimension of the learning concept is decided by the abstraction domain. For the set coverage learning concept in def.~\ref{def:coverage}, the general VC dimension of a single predicate of the form $C_{i} = \{\vx | \omega\vx+ \vb \leq 0\}$ is finite. The actual VC dimension of specific abstract domain and scenario can be easily verified, e.g. the VC dimension of an axis-parallel box is $v=4$. 

\noindent According to Lemma 3.2.3 in~\cite{blumer1989learnability}, the VC dimension of a union set $C$ of convex polygons $C_{i}$ is less than $2vs\log(3s)$, for all $s \geq 1$, where $s$ is the number of sets in $C$. Hence, the required sample size to limit the abstraction error $u \leq \bar{u}$ for the set cover solution $C$ is $\max(\frac{4}{\bar{u}}\log\frac{2}{\delta}, \frac{16vs\log(3s)}{\bar{u}}\log\frac{13}{\bar{u}})$. By underapproximating the number of samples considering conservatively that each concrete state has probability $\underline{p_{QL}}$ of being sampled, we can conclude that an abstraction with misclassification error less than a prescribed $\bar{u}$ can eventually be learned with arbitrary probability $1 - \delta$.

These upper bounds are typically conservatively above the actual number of sampled paths required for most practical scenarios and mainly aim at ensuring the asymptotic convergence of the abstraction process. In sec.~\ref{apxEval}, we will demonstrate empirically on some of the experimental environments how the actual abstraction converges to a prescribed maximum misclassification error for different configuration hyperparameters.

\noindent\emph{Caveats and practical limitations.}
Because our abstract domain uses boxes (with sides parallel to the axes of the state space domain), it is likely to miscalssify safe regions when their boundaries cannot be covered exactly with a union of boxes. In general, a similar argument can be formulated for any finite-accuracy abstract domain. The corner case of using this abstraction is that small safe regions located between two unsafe ones placed at a distance, along any dimension, smaller than the minimum size of a box $d$ may remain misclassified even if the agent happens to sample a concrete point that could discriminate them. In turn, if the optimal policy requires passing through one such small misclassified region, counterexample guidance could reduce the likelihood of the agent exploring it -- but never entirely prevent its exploration. In practice, the problem can be mitigated choosing a smaller value of $d$, or using an abstract domain with a more appropriate performance/accuracy tradeoff for the problem at hand, e.g., octagons or polyhedra.

\vspace{2mm}
\noindent\textbf{Convergence of the online-offline learning process.}
The main aim of offline learning is to discourage the re-exploration of policies that lead to violation of the safety requirement by penalizing the q-values of the involved actions. The penalization of explored unsafe behaviors has the effect of encouraging the exploration of alternative actions in a state, because their q-values ``grow'' relatively to the q-values of the penalized actions and are thus more likely to be selected next. An underlying assumption for the stability of the method is that \emph{there exist an optimal policy that satisfies the safety requirement}. In the following, we will discuss how introducing offline learning does not prevent convergence to such a policy (in fact, as demonstrated experimentally, it accelerates the convergence to it), including sufficient conditions for such convergence to occur. Finally, we will discuss what may happen if all optimal policies violate the safety requirement. Restated from sec.~\ref{secCex}:

\propfour*

The online learning phases are bound to eventually converge to an optimal policy as long as each state-action pair can be visited infinitely often. The offline phases can decrease the relative likelihood of actions involved with policies leading to a violation of the safety requirement, but never prevent their exploration altogether. As a result, when the online phases converge to an optimal policy which satisfies the safety requirement, and assuming the abstract model converged as well, no further offline phases will be triggered (except for possible occasional false positive triggers from the Bayesian hypothesis test). This happened in all our experiments, where occasional offline phases triggered after the agent converged to mainly exploring policies that satisfy the safety requirement could introduce transient fluctuations in the cumulative reward but do not affect convergence to maximum reward in the long run.

\vspace{1mm}
If there exist no maximal-reward policy satisfying the safety requirement, the introduction of offline learning may result in oscillations in the q-values of the actions involved in the maximal-reward policy discovered by the agent. Such oscillations arise from the fact that the maximal-reward policy the online phase converged to is itself, by hypothesis, a counterexample to the safety requirement. In this situation, our method may still reduce the number of failures during exploration by reducing the frequency at which the agent explores the maximal-reward policy, but it will not prevent the agent from converging to such maximal-reward policy in the long run. (Notice that if an alternative policy that achieves the same expected reward while satisfying the safety requirement existed, the introduction of offline learning would encourage its discovery, as discussed in Proposition~\ref{prop4}; however, we are here assuming such policy does not exist.) In this situation, the designer has to accept the need to relax the safety requirement or decide whether to privilege reward over safety, which can be obtained by, e.g., limiting the number of times the same counterexample can be used in an offline phase. This situation is mentioned for the sake of completeness, but it falls outside the scope of this paper, where it is assumed that a maximal-reward policy that satisfies the safety requirement exists and its learning is accelerated via counterexample guidance.

\subsection{Empirical Evaluation of the Abstraction and Sensitivity to Hyperparameters}\label{apxEval}

In this section, we provide supplemental details about the abstraction evaluation, and the sensitivity of hyperparamters in the quality of abstraction and the learning outcomes. We first provide sets of additional hyperparamters in the baseline QL/DQN and our method to ease reproducibility. Then, we illustrate a comprehensive exploration of the concept of incremental abstraction, starting with a concrete example that demonstrates how it functions. This illustration will elucidate the process and its influence on reinforcement learning systems. Next, we analyse the robustness of our learning results with respect to hyperparameter tuning. This analysis will be twofold, focusing on both online and offline related hyperparameters. We aim to present a critical examination of how these parameters affect the quality of the abstraction model and also the overall learning performance. 

\hspace*{\fill} \\

\noindent \textbf{Additional Hyperparamters}
All common hyper-parameters used with QL, DQN and Q-CEX are the same as listed in tab.~\ref{tabHyperparameters}.For DQN specific addtional hyper-parameters, we use the default values given in~\cite{mnih2015human}. For QL-LCRL-specific additional hyper-parameters, we use the same values as~\cite{HasanbeigKA22}. 

\begin{table}[h]\label{tabHyperparameters}
\centering
\begin{tabular}{|l||*{15}{c|}}\hline
\tiny{Environment} &\makebox[2em]{\tiny{\makecell{bayes\\factor}}}&\makebox[3em]{\tiny{\makecell{sim \\ episode}}}&\makebox[3em]{\tiny{\makecell{iteration \\ max}}}&\makebox[4em]{\tiny{\makecell{target up- \\date interval}}}&\makebox[3em]{\tiny{\makecell{max grad \\ norm}}}&\makebox[3em]{\tiny{\makecell{learning \\ starts}}}&\makebox[3em]{\tiny{\makecell{explor\\-ation frac}}}&\makebox[2em]{\tiny{\makecell{grad\\ step}}}&\makebox[2em]{\tiny{\makecell{train \\freq}}}&\makebox[2em]{\tiny{\makecell{batch \\ size}}}\\\hline\hline
\tiny{DiscreteGrid} & \tiny{1}&\tiny{50}&\tiny{5000}&\tiny{-}&\tiny{-}&\tiny{-}&\tiny{-}&\tiny{-}&\tiny{-}&\tiny{-}\\\hline
\tiny{FrozenLake8x8} & \tiny{1}&\tiny{100}&\tiny{199}&\tiny{-}&\tiny{-}&\tiny{-}&\tiny{-}&\tiny{-}&\tiny{-}&\tiny{-}\\\hline
\tiny{HybridGrid} & \tiny{1}&\tiny{50}&\tiny{5000}&\tiny{1000}&\tiny{1}&\tiny{10000}&\tiny{0.2}&\tiny{128}&\tiny{100}&\tiny{128}\\\hline
\tiny{MarsRover} & \tiny{1}&\tiny{50}&\tiny{3000}&\tiny{1000}&\tiny{0.1}&\tiny{1000}&\tiny{0.3}&\tiny{128}&\tiny{100}&\tiny{128}\\\hline
\end{tabular}
\caption{Additional Hyperparamters in QL/DQN and Q-CEX}
\label{tab:hyperparamters}
\end{table}

\begin{figure}\label{fig:incrementalabs}
\centering    
\subfigure[Initial $M_a$] {
 \label{fig: incremental_ab1}     
\includegraphics[width=0.25\columnwidth]{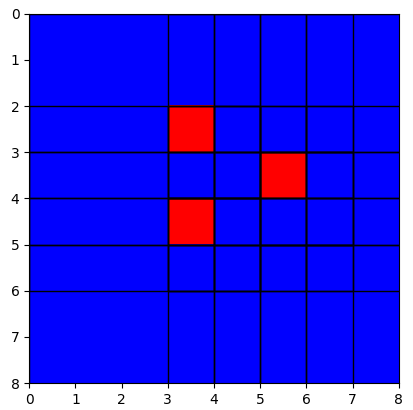}  
}\quad \quad
\subfigure[Refined $M_a$] { 
\label{fig: incremental_ab2}     
\includegraphics[width=0.25\columnwidth]{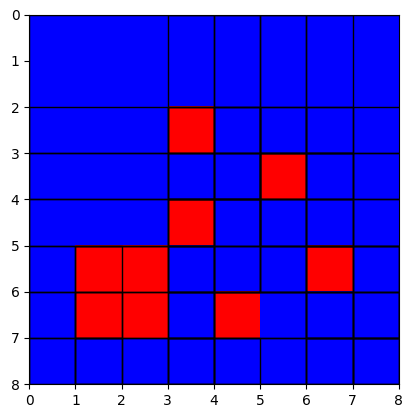}   
}\quad \quad
\subfigure[Concrete Layout] { 
\label{fig: incremental_ab0}     
\includegraphics[width=0.24\columnwidth]{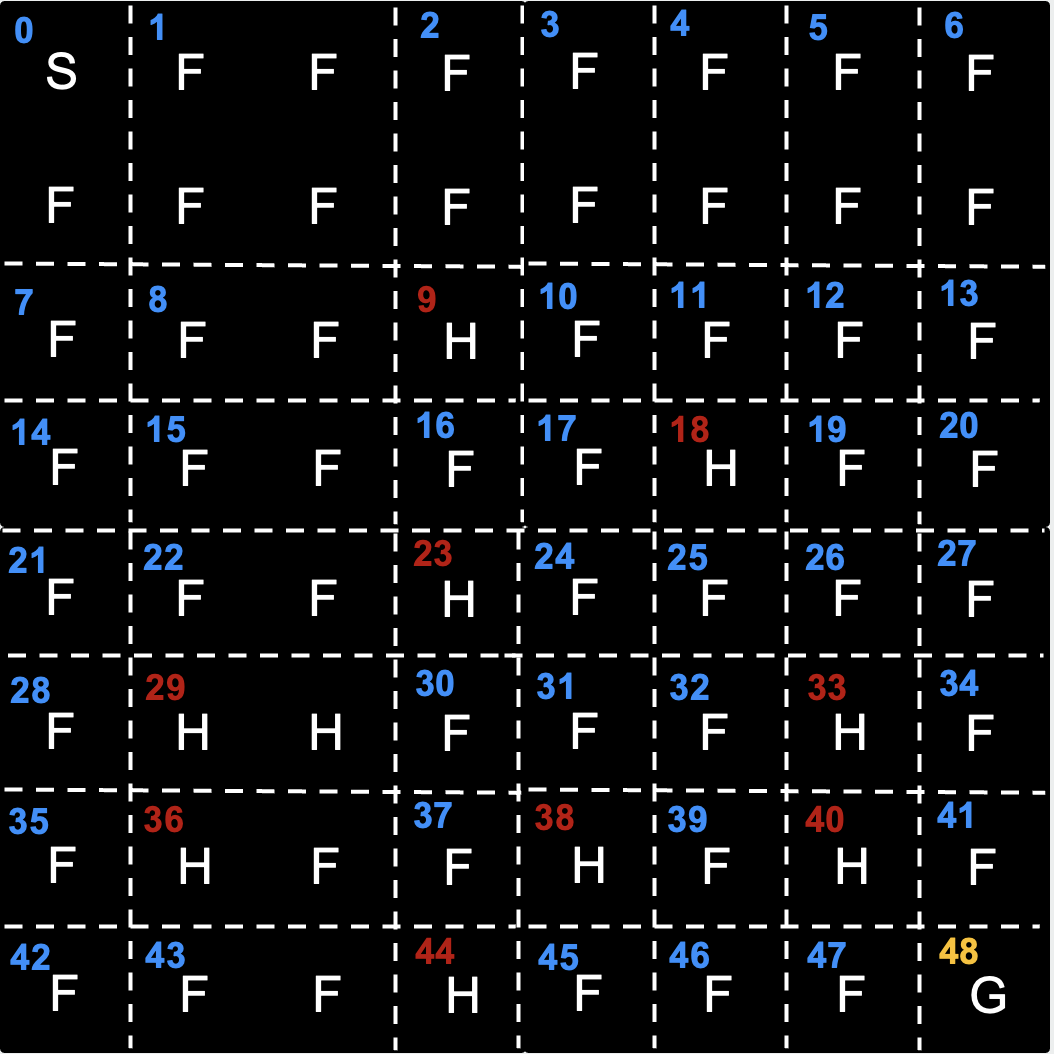}   
}
\caption{Incremental abstraction Demonstration in Frozenlake8x8}
\end{figure}

\noindent \textbf{Incremental Geometric Abstraction.} When new unsafe states are discovered during online exploration, which were previously abstracted as safe, the abstract state space is updated incrementally using a branch-and-bound strategy to separate the new unsafe point. We demonstrate
this process taking as example the Frozenlake8x8 environment in fig.~\ref{fig:incrementalabs}. The initial abstract MDP, used for counterexample generation, after the first 50 online episodes is shown in fig.\ref{fig: incremental_ab1}. With the increasing number of explored points, more safety-relevant information can be acquired, and we employ the branch and bound method incrementally to cover the newly discovered unsafe states and refine the safe states adjacent to the newly discovered unsafe regions, as shown in fig.\ref{fig: incremental_ab2}. The ground-truth layout is included in fig.~\ref{fig: incremental_ab0} for reference, where \texttt{H} indicates unsafe regions. 

\noindent \textbf{Robustness of CEX-guided RL against hyperparamter tuning.} We assessed the robustness of our proposed method in terms of hyperparameter tuning, focusing on two main aspects: the quality of the resulting abstraction model and the learning performance, measured by accumulated safety rates and average rolling rewards.

\begin{figure}[!h] 
\centering    
\subfigure[fpr=0.35, $\epsilon$=0.1] {
\includegraphics[width=0.25\columnwidth]{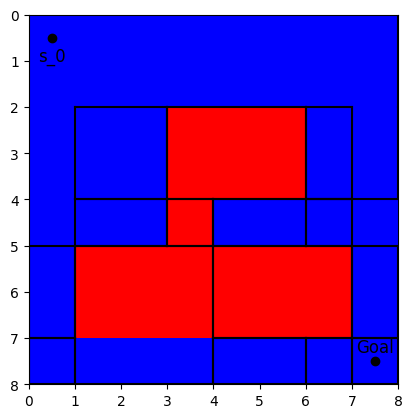}  
}\quad
\subfigure[fpr=0.35, $\epsilon$=0.05] { 
\includegraphics[width=0.25\columnwidth]{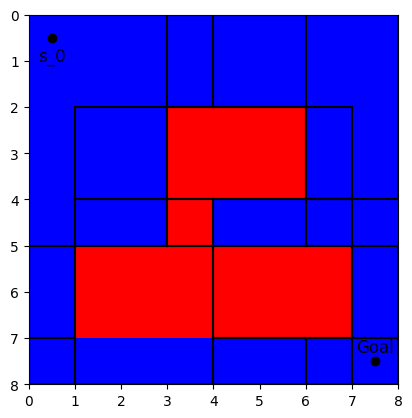}     
}
\subfigure[fpr=0.35, $\epsilon$=0.01] {
\includegraphics[width=0.25\columnwidth]{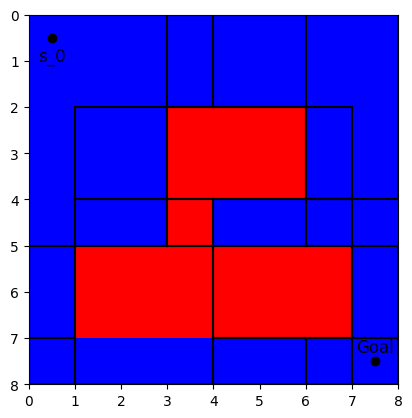}  
}\quad
\subfigure[fpr=0.15, $\epsilon$=0.1] { 
\includegraphics[width=0.25\columnwidth]{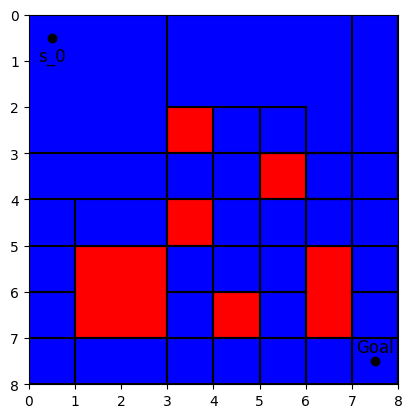}     
}  
\subfigure[fpr=0.15, $\epsilon$=0.05] {
\includegraphics[width=0.25\columnwidth]{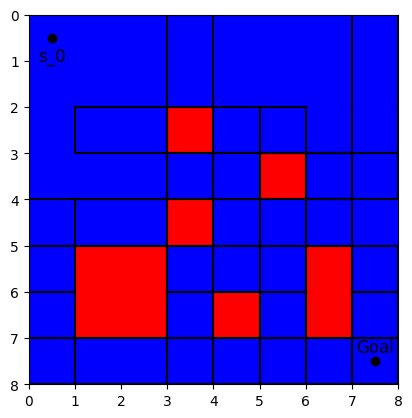} 
}\quad
\subfigure[fpr=0.15, $\epsilon$=0.01] { 
\includegraphics[width=0.25\columnwidth]{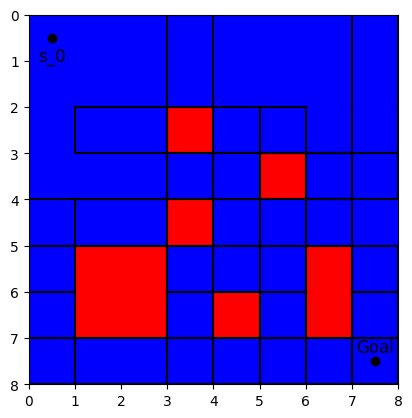}     
}
\subfigure[fpr=0.01, $\epsilon$=0.1] { 
\includegraphics[width=0.25\columnwidth]{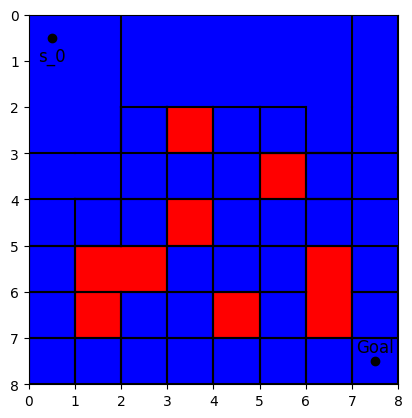}     
}  
\subfigure[fpr=0.01, $\epsilon$=0.05] {
\includegraphics[width=0.25\columnwidth]{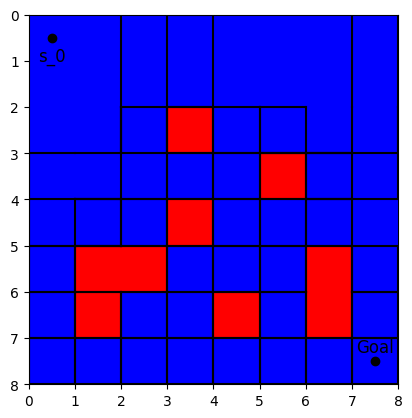}  
}\quad
\subfigure[fpr=0.01, $\epsilon$=0.01] { 
\includegraphics[width=0.25\columnwidth]{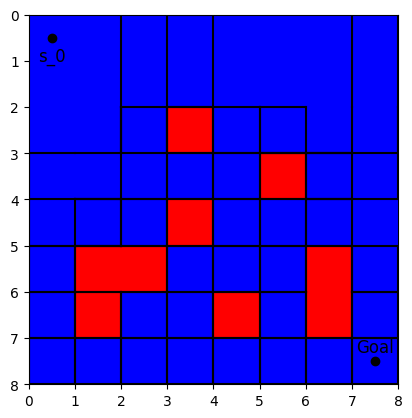}     
} 
\caption{Abstraction of FrozenLake8x8 with different fpr and $\epsilon$}  
\label{fig:abstraction_frozenlake}
\end{figure}

\begin{figure}[h] 
\centering    
\subfigure[fpr=0.35, $\epsilon$=0.1] {
\includegraphics[width=0.25\columnwidth]{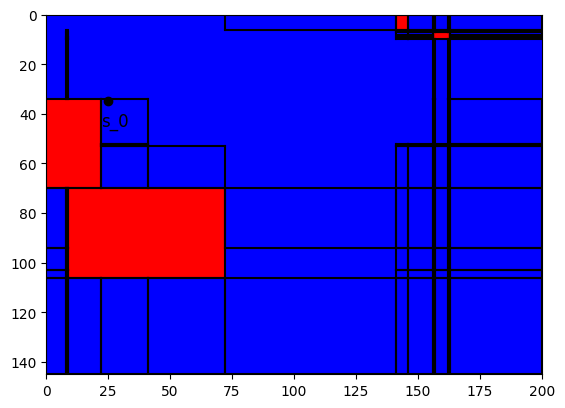}  
}\quad
\subfigure[fpr=0.35, $\epsilon$=0.05] { 
\includegraphics[width=0.25\columnwidth]{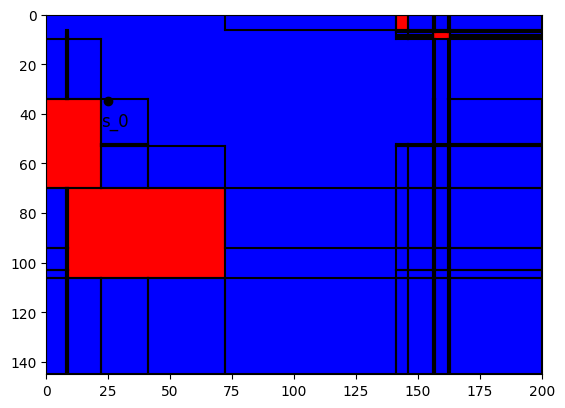}     
}
\subfigure[fpr=0.35, $\epsilon$=0.01] {
\includegraphics[width=0.25\columnwidth]{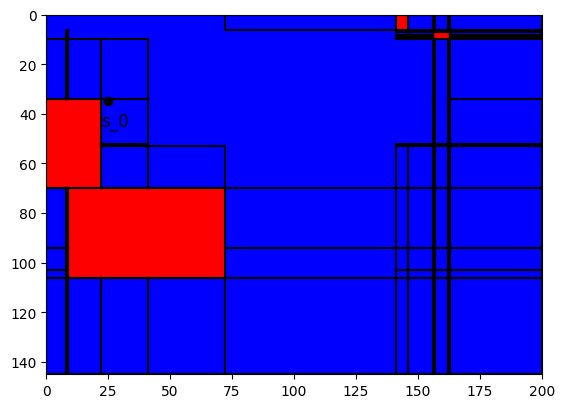}  
}\quad
\subfigure[fpr=0.25, $\epsilon$=0.1] { 
\includegraphics[width=0.25\columnwidth]{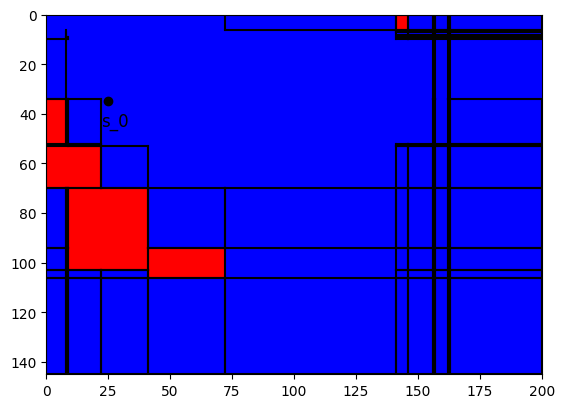}     
}  
\subfigure[fpr=0.25, $\epsilon$=0.05] {
\includegraphics[width=0.25\columnwidth]{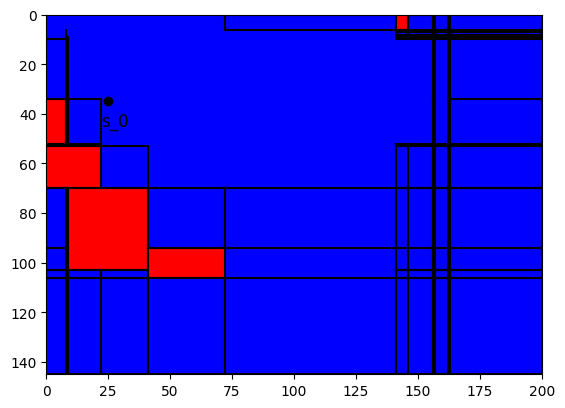} 
}\quad
\subfigure[fpr=0.25, $\epsilon$=0.01] { 
\includegraphics[width=0.25\columnwidth]{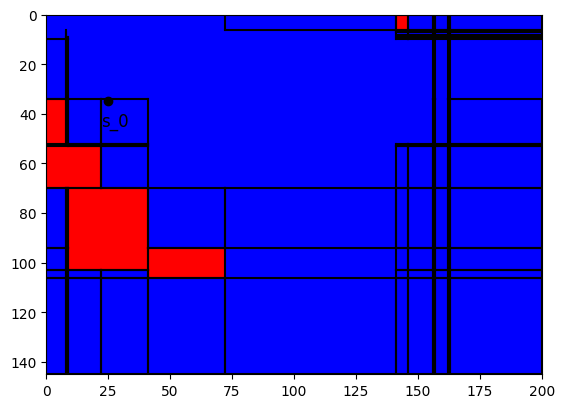}     
}
\subfigure[fpr=0.10, $\epsilon$=0.1] { 
\includegraphics[width=0.25\columnwidth]{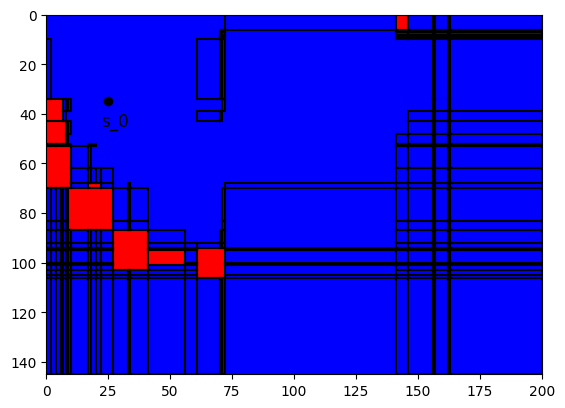}     
}  
\subfigure[fpr=0.10, $\epsilon$=0.05] {
\includegraphics[width=0.25\columnwidth]{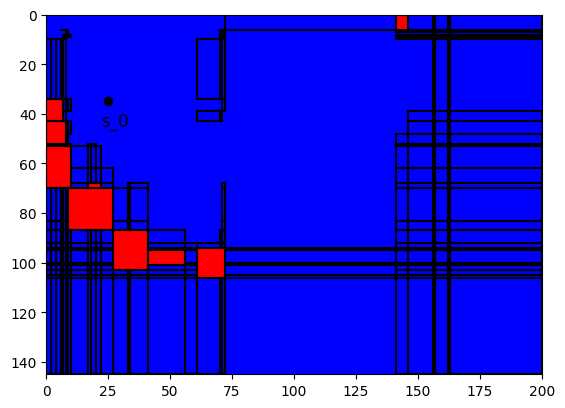}  
}\quad
\subfigure[fpr=0.10, $\epsilon$=0.01] { 
\includegraphics[width=0.25\columnwidth]{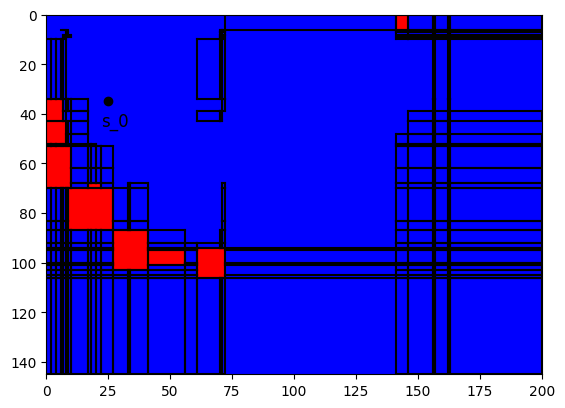}     
}  
\caption{Abstraction Model of Marsrover with different fpr and $\epsilon$}
\label{fig:abstraction_marsrover}
\end{figure}

To evaluate simulation model quality, we present results using varying false positive rates (FPR) and $\epsilon$ for $\epsilon$-bisimulation merging in both Frozenlake8x8 in fig~\ref{fig:abstraction_frozenlake} and MarsRover environments in fig.~\ref{fig:abstraction_marsrover}. A lower FPR results in a more precise abstraction, which includes less safe concrete states from the unsafe abstract states (in environments like MarsRover, safe concrete states cannot be completely excluded with boxes). 

Smaller $\epsilon$ values lead to less merging of similar states in the abstract state space. We recall the hierarchical $\epsilon$-simulation concept, demonstrating that its inherent structure provides a beneficial trade-off between computational complexity and accuracy in proximity to unsafe states. Because the merge operation is constrained to states with identical labels and follows a hierarchical fashion (i.e., states from lower levels merge into higher ones, with transition probabilities at the higher level normalised by the sum from merged lower-level states), a lesser degree of merge due to $\epsilon$-simulation is anticipated in the vicinity of unsafe states. 

\noindent This hierarchical process indirectly optimises the resolution in the state space where it is most crucial, specifically near unsafe states, while permitting a coarser merge in safer regions. Such non-uniform merging fosters a more efficient balance between the computational complexity of the counterexample generation optimisation problem, which depends on the number of states in the abstraction, and the effectiveness of the counterexamples in accurately selecting concrete transitions from the agent's past experience near unsafe concrete regions.

\noindent Intuitively, given a specific $\epsilon$, the final minimised abstraction is expected to be more precise near states labeled as unsafe due to the merge operation's restriction to adjacent abstract states. In these areas, a limited number of steps and corresponding traversal of adjacent states suffices to differentiate the trajectory outcome under consideration. In contrast, a coarser abstraction is produced when more steps are required to determine the trajectory's outcome. This characteristic of the hierarchical $\epsilon$-simulation provides a tailored abstraction mechanism that varies with the safety of the region, enhancing the effectiveness of the abstract model.

Regarding learning performance, we demonstrate the average accumulated safety rate and the average rolling reward are robust under different hyperparameters, by varying the penalty and number of offline episodes for each simulation model during offline learning phase, and varying the Bayes factor and safety check interval during the online exploration in tab.~\ref{tblHyperparameters}. 

\noindent To estimate the probability of safety requirement violation and trigger offline learning when necessary, we utilise the Bayesian hypothesis testing estimator~\cite{filieri2014statistical} with the Bayes factor as a hyperparamter. This estimator can be initialised with a minimum number of samples prior to accepting its decision, and then determines whether the safety requirements are satisfied or violated based on the number of samples collected during the most recent online learning session. If the likelihood of violating the safety requirement significantly surpasses that of satisfying it, the offline learning phase is subsequently initiated.

\noindent Within the online learning session, we assess the Bayes factor of these two hypotheses, represented as $\frac{P(H_{0}|S)}{P(H_{1}|S)}$, using the real-time experience dataset. The magnitude of the given Bayes factor decides the frequency of triggering offline learning. This frequency setting engenders a trade-off between safety and computational cost: higher frequency leads to increased computational cost. A high-frequency setting may also induce over-conservative learning performance. This happens when excessive offline learning guidance inadvertently penalises the learning process that entails a reasonable level of risk. In corner cases with very small Bayes factors, the agent may opt to explore solely safe regions, failing to achieve the learning objective.

\begin{table}[h]
\centering
\begin{tabular}{|l||*{3}{c|}}\hline
\backslashbox[50mm]{Penalty}{\scriptsize{SimTrainEpisodes}}
&\makebox[3em]{20}&\makebox[3em]{50}&\makebox[3em]{100}\\\hline\hline
-0.1 &$43.05\% / 0.595$& $45.5\% / 0.59$& $44.35\% / 0.59$\\\hline
-0.5 &$42.26\% /0.614$& $43.05\% / 0.67$& $44.8\% / 0.57$\\\hline
-1 &$44.5\% / 0.592$& $40.4\% / 0.549$& $45.8\% / 0.619$\\\hline
\end{tabular}

\begin{tabular}{|l||*{3}{c|}}\hline
\backslashbox[50mm]{Bayes Factor}{\scriptsize{SafetyCheckInterval}}
&\makebox[3em]{500}&\makebox[3em]{1000}&\makebox[3em]{2000}\\\hline\hline
1 &$41.5\% / 0.5$& $44.35 \% / 0.59$& $42.02\% / 0.655$\\\hline
2 &$40.73\% / 0.554$& $38.71\% / 0.532$& $38.2\% / 0.542$\\\hline
10 &$39.6\% / 0.708$& $37.71\% / 0.635$& $36.8\% / 0.521$\\\hline
\end{tabular}
\caption{Robustness of Cex-Guided RL against hyperparamter tuning in slippery FrozenLake8x8}
\label{tblHyperparameters}
\end{table}

 \end{document}